\documentclass[sigconf]{acmart}

\usepackage{graphicx}
\usepackage{algorithm}
\usepackage{algorithmic}
\usepackage{subfigure}
\usepackage{array}
\usepackage{setspace}
\usepackage{float}
\usepackage{url}
\usepackage{balance}
\usepackage{amsfonts}
\usepackage{amsmath}
\usepackage{rotating}
\usepackage{multirow}
\usepackage{color}
\usepackage{bm}
\usepackage{enumitem}
\usepackage[T1]{fontenc}
\usepackage{caption}
\newcommand{\bml}{{\bm l}}
\newcommand{\bmmu}{{\bm \mu}}
\newcommand{\bmx}{{\bm x}}
\newcommand{\bmd}{{\bm d}}
\newcommand{\bmw}{{\bm w}}
\newcommand{\bmm}{{\bm m}}
\newcommand{\bmz}{{\bm z}}
\newcommand{\bmth}{{\bm \theta}}

\copyrightyear{2021}
\acmYear{2021}
\setcopyright{acmcopyright}\acmConference[WSDM '21]{Proceedings of the Fourteenth ACM International Conference on Web Search and Data Mining}{March 8--12, 2021}{Virtual Event, Israel}
\acmBooktitle{Proceedings of the Fourteenth ACM International Conference on Web Search and Data Mining (WSDM '21), March 8--12, 2021, Virtual Event, Israel}
\acmPrice{15.00}
\acmDOI{10.1145/3437963.3441730}
\acmISBN{978-1-4503-8297-7/21/03}

\begin{document}
\title{Hierarchical Metadata-Aware Document Categorization \\ under Weak Supervision} 
\author{Yu Zhang$^{1}$, Xiusi Chen$^{2}$, Yu Meng$^{1}$, Jiawei Han$^{1}$}
\affiliation{
\institution{$^1$Department of Computer Science, University of Illinois at Urbana-Champaign, IL, USA} 
\institution{$^2$Department of Computer Science, University of California, Los Angeles, CA, USA}
\institution{$^{1}$\{yuz9, yumeng5, hanj\}@illinois.edu, \ \ \ $^2$xchen@cs.ucla.edu}
}

\begin{abstract}
Categorizing documents into a given label hierarchy is intuitively appealing due to the ubiquity of hierarchical topic structures in massive text corpora. Although related studies have achieved satisfying performance in fully supervised hierarchical document classification, they usually require massive human-annotated training data and only utilize text information. However, in many domains, (1) annotations are quite expensive where \textit{very few training samples} can be acquired; (2) documents are accompanied by \textit{metadata} information. Hence, this paper studies how to integrate the label hierarchy, metadata, and text signals for document categorization under weak supervision. We develop \textsc{HiMeCat}, an embedding-based generative framework for our task. Specifically, we propose a novel \textit{joint representation learning module} that allows simultaneous modeling of category dependencies, metadata information and textual semantics, and we introduce a \textit{data augmentation module} that hierarchically synthesizes training documents to complement the original, small-scale training set. Our experiments demonstrate a consistent improvement of \textsc{HiMeCat} over competitive baselines and validate the contribution of our representation learning and data augmentation modules. 
\end{abstract}

% \begin{CCSXML}
% 	<ccs2012>
% 	<concept>
% 	<concept_id>10002951.10003317.10003347.10003356</concept_id>
% 	<concept_desc>Information systems~Clustering and classification</concept_desc>
% 	<concept_significance>500</concept_significance>
% 	</concept>
% 	<concept>
% 	<concept_id>10010147.10010257</concept_id>
% 	<concept_desc>Computing methodologies~Machine learning</concept_desc>
% 	<concept_significance>500</concept_significance>
% 	</concept>
% 	</ccs2012>
% \end{CCSXML}

% \ccsdesc[500]{Information systems~Clustering and classification}
% \ccsdesc[500]{Computing methodologies~Machine learning}

% \keywords{}

\maketitle

\begin{spacing}{0.97}
\section{Introduction}
\begin{figure}[t]
    \centering
    \subfigure[\textbf{GitHub Repository.} Label Hierarchy: PaperWithCode Task Taxonomy (\url{https://paperswithcode.com/sota}); Text: Description and README; Metadata: User and Tag.]{
    \includegraphics[width=0.99\linewidth]{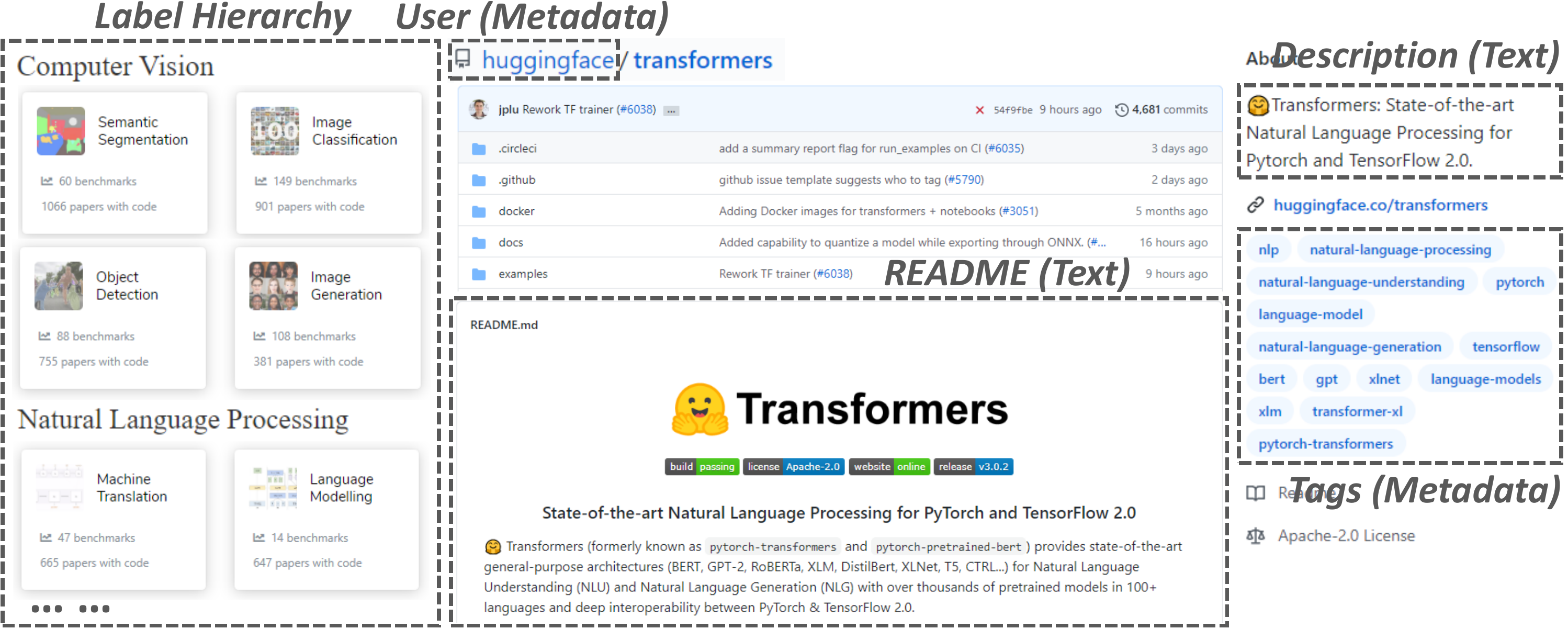}} \\
    \vspace{-0.5em}
    \subfigure[\textbf{arXiv Paper.} Label Hierarchy: arXiv Category Taxonomy (\url{https://arxiv.org/category_taxonomy}); Text: Title and Abstract; Metadata: Author.]{
    \includegraphics[width=0.99\linewidth]{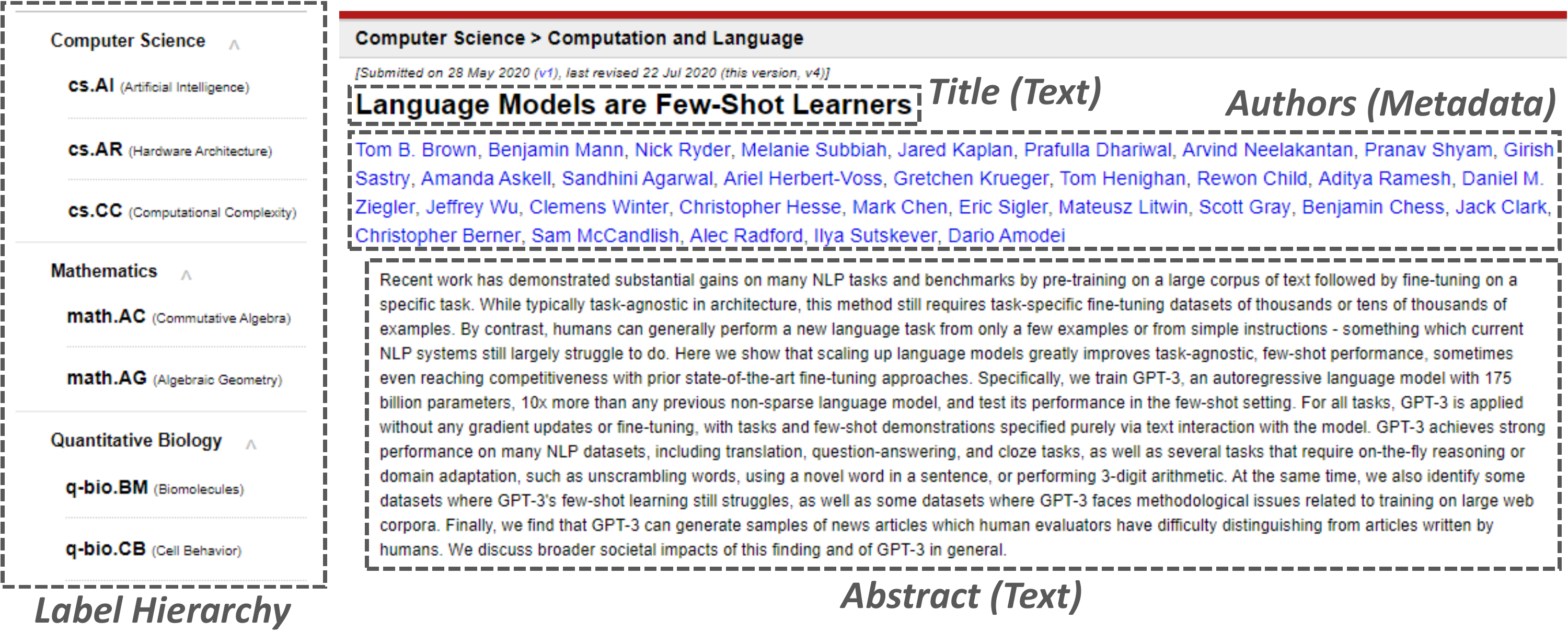}} \\
    \vspace{-0.5em}
    \subfigure[\textbf{Amazon Review.} Label Hierarchy: Amazon Product Catalog \cite{mao2020octet}; Text: Title and Review; Metadata: User and Product.]{
    \includegraphics[width=0.99\linewidth]{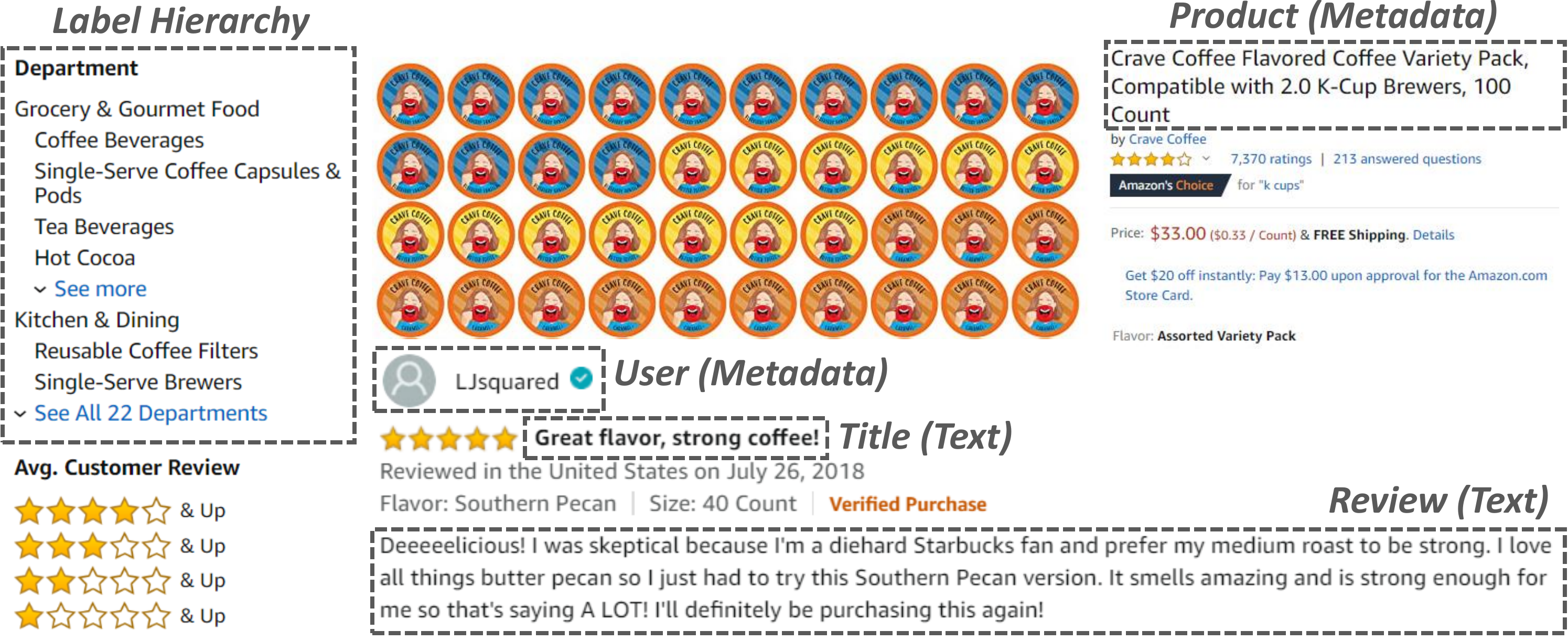}} \\
    \vspace{-1em}
    \caption{Three examples of documents with metadata and corresponding label hierarchies.} 
    \vspace{-1.5em}
    \label{fig:example}
\end{figure}

Document categorization is an important task in text mining, with many real applications such as sentiment analysis \cite{tang2015learning}, location prediction \cite{chen2020context} and scientific paper tagging \cite{li2019hiercon}. Given a large text corpus, automatically inferring the category of each document not only enables effective organization of the data, but also benefits downstream tasks like document retrieval \cite{wu2001using}. Hierarchical document categorization \cite{dumais2000hierarchical,gopal2013recursive} further considers relationships among categories and classifies documents into a given label hierarchy. Leveraging such hierarchical structures is shown to be effective and necessary due to their pervasiveness in web directories\footnote{\url{https://en.wikipedia.org/wiki/Yahoo!_Directory}}, business category lists\footnote{\url{https://www.yelp.com/developers/documentation/v3/all_category_list}} and many other domains (Figure \ref{fig:example}).

Although recent studies on hierarchy-aware neural models \cite{wehrmann2018hierarchical,mao2019hierarchical,zhou2020hierarchy} and pre-trained language models \cite{devlin2019bert,sun2019fine} greatly improve the performance of (hierarchical) text categorization, they are less concerned with two challenges in real applications:

\vspace{1mm}

\noindent \textbf{Limited Training Data.} Neural classifiers are data hungry and require a significant amount of manually labeled training documents to achieve good performance. In some scientific domains (e.g., arXiv papers and GitHub repositories), to obtain a large training set is expensive because annotations have to be acquired from domain experts. In these scenarios, it would be favorable to perform classification without much annotation effort. To be specific, one may only expect very few (e.g., 5) training examples for each category. 

\vspace{1mm}

\noindent \textbf{Heterogeneous Signals.} Documents on the Web are often accompanied by \textit{metadata} information. We raise three examples in Figure \ref{fig:example}: each GitHub repository has its creator and several topic tags; each arXiv paper has its author(s); each Amazon review is associated with a user and a product. These signals, as potential category indicators, should be considered together with the plain text sequence. Moreover, the \textit{label hierarchy}, which reflects category dependencies and correlation, should also be viewed as one type of signal. In recognition of these heterogeneous signals, how can we jointly leverage them during category inference?

In this paper, to tackle the two challenges, we develop \textsc{HiMeCat}, an embedding-based generative framework for \textsc{Hi}erarchical \textsc{Me}tadata-aware document \textsc{Cat}egorization under weak supervision. 
\textsc{HiMeCat} features a hierarchical generative process in the embedding space
% , where (1) \textit{label} embeddings are generated layer by layer according to the label hierarchy, and the embedding of a child label depends on its parent; (2) \textit{document} embeddings are generated based on label and \textit{metadata} representations; (3) for each document, a sequence of \textit{words} are then generated given the document embedding.
to simultaneously model (1) \emph{category dependencies} via a top-down generative assumption along the label hierarchy, (2) \emph{metadata information} via a joint generative assumption that documents are dependent on both their categories and metadata, and (3) \emph{textual semantics} via a document-word generative assumption.
This hierarchical generative process guides the two important steps of our method: (1) \emph{representation learning}: Maximizing the likelihood of the hierarchical generative process yields an embedding learning objective which jointly optimizes the representations of the label hierarchy, metadata and texts to effectively exploit the \textbf{heterogeneous signals};
% to fully exploit \textbf{heterogeneous signals}, we propose an embedding module to jointly learn the representations of label hierarchy, metadata, documents and words, and we show that learning the embedding is essentially maximizing the log-likelihood of the hierarchical generative process; 
(2) \emph{data augmentation}: Following the hierarchical generative process allows us to synthesize training documents that augment the original, small-scale training set, which overcomes the \textbf{limited training data} challenge.
% to address the bottleneck of \textbf{limited training data}, we generate synthesized training samples to augment the original training data. Specifically, given a category in the label hierarchy, we follow the aforementioned steps in the hierarchical generative process to synthesize documents as sequences of words.
Combining real and synthesized training data, we train a neural classifier for hierarchical text categorization by taking word representations learned from the previous step as embedding initialization.

Inspired by related studies on spherical hierarchical clustering \cite{gopal2014mises} and spherical text embedding \cite{meng2019spherical} which better capture directional similarities and outperform Euclidean space models, we propose to define our hierarchical generative process in the spherical space, where conditional probabilities are described by the von Mises-Fisher (vMF) distribution \cite{fisher1953dispersion} and Riemannian optimization \cite{bonnabel2013stochastic} is adopted for embedding learning.
% Then, in our generation module, we also use the vMF distribution when synthesizing documents given label embeddings.

To summarize, our contributions are as follows:
\begin{itemize}[leftmargin=*]
\item We explore the task of jointly leveraging limited supervision and metadata information for hierarchical text classification.
 
\item We develop \textsc{HiMeCat} with a representation learning module and a data augmentation module guided by a novel hierarchical generative process in the spherical space. Our representation learning module jointly learns embeddings from heterogeneous signals, and our data augmentation module synthesizes training documents to address the supervision scarcity bottleneck.

\item We conduct comprehensive experiments on three datasets from different domains and observe a consistent improvement of \textsc{HiMeCat} over competitive baselines. We also show that leveraging the hierarchy, leveraging metadata, and generating training samples are all beneficial to the categorization performance.
%\footnote{Code and datasets will be released upon the acceptance of the paper.}
\end{itemize}

\section{Preliminaries}
\subsection{Problem Definition}
The inputs of our hierarchical metadata-aware text categorization task are a collection of documents $\mathcal{D} = \{d_1,...,d_{|\mathcal{D}|}\}$ and a label hierarchy $\mathcal{T}$. Each document $d_i \in \mathcal{D}$ has both text and metadata information. We explain these concepts one by one.

\vspace{1mm}

\noindent \textbf{Text.} In this paper, text refers to all free-text fields of a document. For example, in Figure \ref{fig:example}, a GitHub repository has its description and README file as text information; an arXiv paper has its title and abstract. To simplify our discussion, we concatenate all these fields into one word sequence, denoted as $w_1w_2...w_n$.

\vspace{1mm}

\noindent \textbf{Metadata.} Each document can have multiple types of metadata (e.g., an Amazon review has both user and product information). Given a specific metadata type, there may be an arbitrary number (could be zero) of metadata instances associated with a document (e.g., a repository may have no tags, and a paper can have more than one author). For a document $d$, we represent its metadata as a set of metadata instances $\mathcal{M}_d = \{m_1,...,m_{|\mathcal{M}_d|}\}$.

\vspace{1mm}

\noindent \textbf{Label Hierarchy.} We assume the label hierarchy $\mathcal{T}$ has a tree structure. Each node $l \in \mathcal{T}$ represents a category. If, in some cases, $\mathcal{T}$ is a directed acyclic graph (DAG), we follow \cite{zhou2020hierarchy} and convert $\mathcal{T}$ to a tree by distinguishing each label node as a single-path node.

Formally, we define our task as follows:

\vspace{1mm}

\noindent \textbf{Problem Definition.} Given a collection of documents $\mathcal{D}$ and a label hierarchy $\mathcal{T}$, where each \textit{leaf} category $l \in \mathcal{T}$ is characterized by a small set of training documents $\mathcal{D}_l \subset \mathcal{D}$, the task is to assign appropriate category labels to the remaining documents $\mathcal{D}\backslash (\cup_l \mathcal{D}_l)$. The labels of each document should form a path in $\mathcal{T}$.

\subsection{The Von Mises-Fisher Distribution}
\label{sec:vmf}
The von Mises-Fisher (vMF) distribution \cite{fisher1953dispersion,mardia2009directional} will be extensively used in our proposed framework. It defines a probability density over a unit sphere and is parameterized by a mean direction vector $\bmmu$ and a concentration parameter $\kappa$. Formally, let $\mathbb{S}^{p-1} = \{\bmx \in \mathbb{R}^p:||\bmx||_2=1\}$ denote the $p$-dimensional unit sphere. The probability density function for $\bmx \in \mathbb{S}^{p-1}$, $||\bmmu||_2=1$, $\kappa > 0$ is given by
\begin{equation}
f_{\rm vMF}(\bmx|\bmmu,\kappa) = c_p(\kappa)\exp(\kappa \cdot \cos(\bmx,\bmmu)).
\label{eqn:vmf}
\end{equation}
Here, the normalization constant $c_p(\kappa)$ is 
\begin{equation}
c_p(\kappa) = \frac{\kappa^{p/2-1}}{(2\pi)^{p/2}I_{p/2-1}(\kappa)},
\end{equation}
where $I_r(\cdot)$ represents the modified Bessel function of the first kind at order $r$ \cite{fisher1953dispersion,mardia2009directional}.
Intuitively, the vMF distribution is an analogue of the Gaussian distribution on a sphere. The distribution concentrates around the mean direction $\bmmu$, and is more concentrated if $\kappa$ is large.

\section{Model}
\subsection{A Hierarchical Generative Process}
\label{sec:process}
\begin{figure}[t]
    \centering
    \subfigure[The Hierarchical Generative Process]{
    \includegraphics[height=4.0cm]{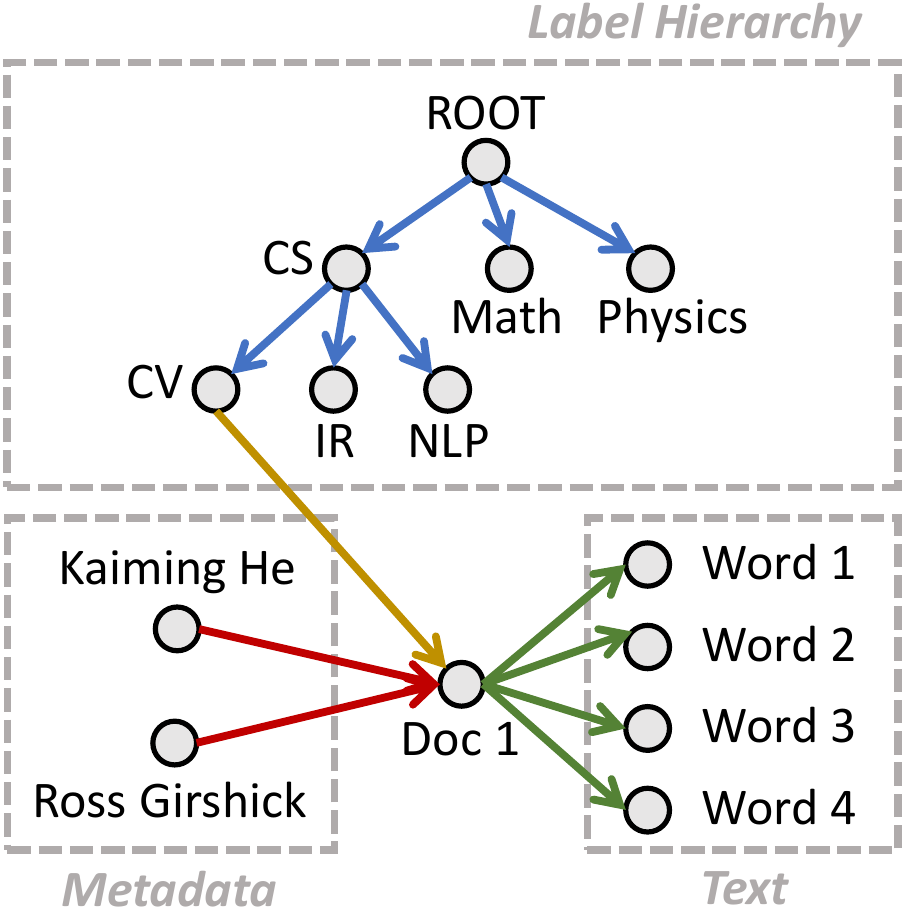}}
    \hspace{2mm}
    \subfigure[The Joint Spherical Embedding Space]{
    \includegraphics[height=4.0cm]{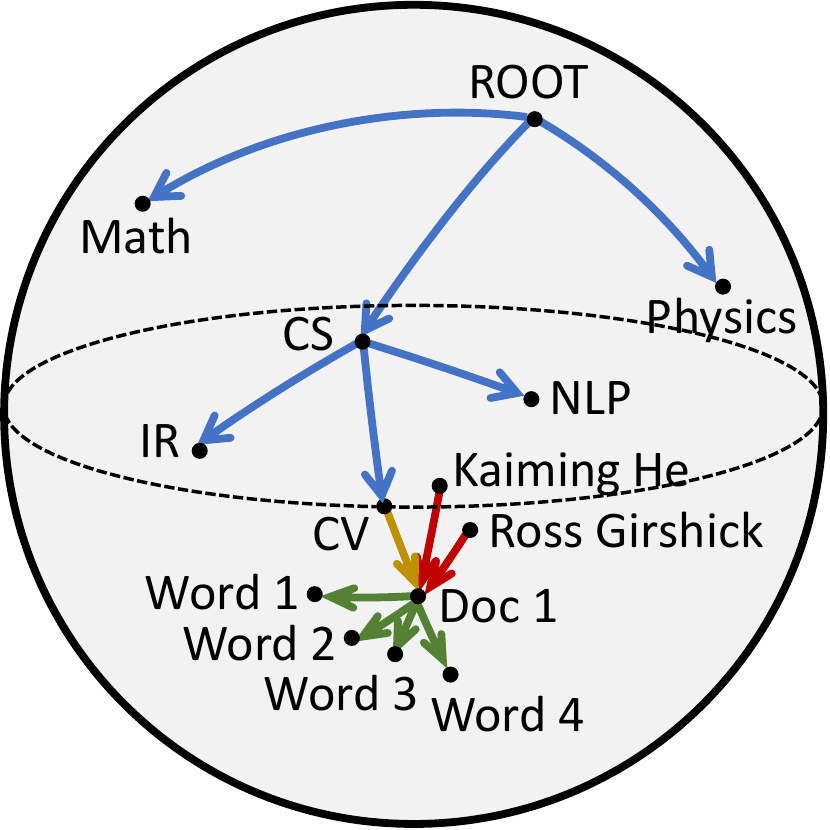}} \\
    \vspace{-1em}
    \caption{Illustration of the proposed hierarchical generative process and the joint hierarchy-metadata-text representation learning module guided by this process.} 
    \vspace{-1em}
    \label{fig:story}
\end{figure}

\noindent \textbf{Motivation.} In recognition of various types of information (i.e., the label hierarchy, weak supervision, metadata, and text), we propose a probabilistic generative process to characterize these heterogeneous signals. 
By adopting the vMF distribution, our generative process is defined in a spherical space (i.e., $\mathbb{S}^{p-1}$) instead of a Euclidean one (i.e., $\mathbb{R}^{p}$). The motivation of doing so is inspired by two lines of related studies. First, the vMF distribution has demonstrated its effectiveness in several tasks including hierarchical clustering \cite{gopal2014mises} and text sequence generation \cite{kumar2018mises}, which bears some similarities with our hierarchical representation learning and data augmentation modules, respectively. Second, normalizing word embedding vectors onto a sphere \cite{levy2015improving,xing2015normalized} or directly training embeddings in a spherical space \cite{meng2019spherical} can better capture directional similarities, which is proved to be beneficial to tasks like semantic similarity search and document classification \cite{meng2019spherical}.

As shown in Figure \ref{fig:story}(a), the process can be divided into the following steps.

\vspace{1mm}

\noindent \textbf{Parent Label $\rightarrow$ Child Label.} Given the label hierarchy $\mathcal{T}$, the semantics of each child category $l_c$ should largely depend on the semantics of its parent category $l_p$. In other words, if we use an embedding vector to represent each category, the child embedding $\bml_c$ should be close to its parent embedding $\bml_p$. Inspired by the softmax function used in word embedding \cite{mikolov2013distributed} and network embedding \cite{tang2015line}, we define the generation probability as
\begin{equation}
    p(l_c|l_p) \propto \exp(\bml_c^T\bml_p).
\label{eqn:pc}
\end{equation}
Here, one needs to notice that $p(\cdot|l_p)$ describes a probability distribution of an embedding vector $\bml_c$. Since $\bml_c$ can be any point in a \textit{continuous} embedding space, $p(\cdot|l_p)$ should also be a \textit{continuous} distribution. Therefore, we cannot fully borrow the softmax function which defines a \textit{discrete} choice from finite candidates. Instead, the vMF distribution introduced in Section \ref{sec:vmf} is a proper tool here. 

Assume all category embedding vectors reside on a sphere $\mathbb{S}^{p-1}$. Recall Eq. (\ref{eqn:vmf}). If we set $\kappa=1$, then
\begin{equation}
f_{\rm vMF}(\bmx|\bmmu,1) = c_p(1)\exp(\cos(\bmx,\bmmu)) = c_p(1)\exp(\bmx^T\bmmu).
\label{eqn:vmf1}
\end{equation}
By comparing Eq. (\ref{eqn:vmf1}) with Eq. (\ref{eqn:pc}), we know that the vMF distribution meets our requirement of the generation probability (because $c_p(1)$ is a constant given the dimension $p$). Thus, we rewrite Eq. (\ref{eqn:pc}) as 
\begin{equation}
    p(l_c|l_p) = f_{\rm vMF}(\bml_c|\bml_p,1) = c_p(1)\exp(\bml_c^T\bml_p),\ \ \ \forall \bml_c \in \mathbb{S}^{p-1}.
\label{eqn:pc1}
\end{equation}

%Despite the success of these hyperbolic models in their tasks (e.g., lexical entailment prediction), we argue that they are not the most suitable methods for hierarchical classification tasks. The reason is that they aim to preserve the absolute tree distance. However, from the text classification perspective, documents from lower-level sibling categories are generally more similar than those from higher-level siblings despite the same tree distance.

Given the conditional probability, the category embedding can be generated in a top-down manner: we first determine the embedding of $\mathcal{T}$'s root, and then proceed to the next level to generate embeddings of categories at that level. The embeddings of child labels are supposed to be surrounding the embedding of their parent label, and, according to the property of the vMF distribution, we have $\mathbb{E}[\bml_c|\bml_p] = \bml_p$. This process can be repeated until
we reach the leaf categories.

\vspace{1mm}

\noindent \textbf{Label \& Metadata $\rightarrow$ Document.} Our second step generates document embeddings according to their metadata and label information (if applicable). Consider how we humans write a document given a topic category (or a tag/product). We need to first have a general idea of the article, which can be represented as the document embedding $\bmd$. To impose the coherence between $\bmd$ and its category/metadata information, we assume
\begin{equation}
    p(d|l_d, \mathcal{M}_d) \propto \exp(\bmd^T\bml_d) \cdot \prod_{m_d \in \mathcal{M}_d} \exp(\bmd^T\bmm_d),
\label{eqn:ld}
\end{equation}
where $l_d$ is the category label of $d$ (here, we are referring to the leaf label of $d$); $m_d$ is a metadata instance of $d$; the corresponding embedding variables are in bold. Following the derivation above, by assuming all embedding vectors reside on a sphere, we can rewrite Eq. (\ref{eqn:ld}) as 
\begin{equation}
\begin{split}
p(d|l_d, \mathcal{M}_d) & \propto \prod_{z \in \{l_d\} \cup \mathcal{M}_d} c_p(1)\exp(\bmd^T\bmz) \\
& = \prod_{z \in \{l_d\} \cup \mathcal{M}_d} f_{\rm vMF}(\bmd|\bmz,1),\ \ \ \forall \bmd \in \mathbb{S}^{p-1}.
\label{eqn:ld1}
\end{split}
\end{equation}
One unique problem we need to address here is that label and metadata information can be missing. For example, due to supervision scarcity, only a small proportion of documents in $\mathcal{D}$ have label information. Besides, a GitHub repository may have no tags. In general, suppose $V$ instances $z_1,...,z_V$ are missing in $\{l_d\} \cup \mathcal{M}_d$. In this case, we assume their embeddings $\bmz_1,...,\bmz_V$ can be any vector on the sphere with equal probability (i.e., $\bmz_1,...,\bmz_V \sim U(\mathbb{S}^{p-1})$ i.i.d.). Let $\mathcal{O}_d = \{l_d\} \cup \mathcal{M}_d \backslash \{z_1,...,z_V\}$ be the remaining observable instances. Therefore, Eq. (\ref{eqn:ld1}) becomes 
\begin{equation}
\begin{split}
&\ p(d|l_d, \mathcal{M}_d) \\
\propto &\ \mathbb{E}_{\bmz_1,...,\bmz_V \sim U(\mathbb{S}^{p-1})} \bigg[ \prod_{z \in \{l_d\}\cup \mathcal{M}_d} c_p(1)\exp(\bmd^T\bmz) \bigg] \\
\propto & \int_{\mathbb{S}^{p-1}}...\int_{\mathbb{S}^{p-1}} \bigg(\prod_{z \in \{l_d\}\cup \mathcal{M}_d} c_p(1)\exp(\bmd^T\bmz)\bigg)\ {\rm d}\bmz_1\ ...\ {\rm d}\bmz_V \\
= & \prod_{i=1}^V \bigg(\int_{\mathbb{S}^{p-1}} c_p(1)\exp(\bmd^T\bmz_i)\ {\rm d}\bmz_i\bigg) \cdot \prod_{z \in \mathcal{O}_d} c_p(1)\exp(\bmd^T\bmz) \\
= & \prod_{z \in \mathcal{O}_d} c_p(1)\exp(\bmd^T\bmz) = \prod_{z \in \mathcal{O}_d} f_{\rm vMF}(\bmd|\bmz,1).
\label{eqn:ld2}
\end{split}
\end{equation}
The second-to-last step holds because $\int_{\mathbb{S}^{p-1}} c_p(1)\exp(\bmd^T\bmz_i)\ {\rm d}\bmz_i \equiv 1$ (i.e., the probability density function of a vMF distribution integrates to 1 over the whole sphere). Eq. (\ref{eqn:ld2}) tells us that, when some label/metadata information is missing, we only need to consider the remaining observable part (i.e., $\mathcal{O}_d$) in the conditional probability.

\vspace{1mm}

\noindent \textbf{Document $\rightarrow$ Word.} Based on the document embedding $\bmd$, a sequence of words $w_1,...,w_n$ is then generated to describe the overall semantics of $d$. Similar to Eqs. (\ref{eqn:pc}) and (\ref{eqn:ld}), we assume the probability of word $w_i$ appearing in document $d$ to be
\begin{equation}
    p(w_i|d) \propto \exp(\bmw_i^T\bmd).
\label{eqn:dw}
\end{equation}
Again, we rewrite the probability using a vMF distribution:
\begin{equation}
    p(w_i|d) = f_{\rm vMF}(\bmw_i|\bmd,1) = c_p(1)\exp(\bmw_i^T\bmd),\ \ \ \forall \bmw_i \in \mathbb{S}^{p-1}.
\label{eqn:dw1}
\end{equation}

Eqs. (\ref{eqn:pc1}), (\ref{eqn:ld2}) and (\ref{eqn:dw1}) together fully specify our proposed hierarchical generative process.

\subsection{Joint Representation Learning}
Guided by the hierarchical generative process, we aim to jointly learn the embeddings of all labels, metadata instances, documents and words. Our learning objective can be divided into two parts: one for modeling the label hierarchy information (corresponding to the first step in our generative process), and the other for modeling corpus statistics and metadata information (corresponding to the second and third steps).

\vspace{1mm}

\noindent \textbf{Objective for Label Hierarchy.} Given a parent-child label pair $(l_p, l_c)$, our goal is to maximize the log-likelihood $\log p(l_c|l_p)$ during embedding learning. Inspired by studies on knowledge graph embedding \cite{bordes2013translating,wang2017knowledge}, we adopt the following margin-based ranking objective:
\begin{equation}
    \min\Big(0, \log p(l_c|l_p) - \log p(l_c'|l_p') - \gamma\Big),
\end{equation}
where $\gamma > 0$ is a margin hyperparameter and $(l_p', l_c')$ is a negative training sample. Given the positive pair $(l_p, l_c)$, we generate the negative pair by replacing the parent label with any other label in $\mathcal{T}$ while keeping the child label. 
%The reason we corrupt $l_p$ is because $\forall l_p'\in \mathcal{T} \backslash \{l_p, l_c\}$, $(l_p', l_c)$ is a false parent-child pair (since each child has only one parent in a tree). In contrast, if we corrupt $l_c$, it is still possible that $(l_p, l_c')$ is a true parent-child pair because $l_c'$ could be a sibling of $l_c$. 
According to this corruption strategy, our objective can be defined as
\begin{equation}
  \mathcal{J}_L = \sum_{l_c \in \mathcal{T}\backslash \{{\rm ROOT}\}} \sum_{l_p' \in \mathcal{T} \backslash \{l_p, l_c\}} \min\Big(0, \log p(l_c|l_p) - \log p(l_c|l_p') - \gamma_L \Big).
\end{equation}
Based on the definition of $p(l_c|l_p)$ in Eq. (\ref{eqn:pc1}), we have
\begin{equation}
\begin{split}
 &\ \log p(l_c|l_p) - \log p(l_c|l_p') \\ 
 = &\ \log \Big(c_p(1)\exp(\bml_c^T\bml_p)\Big) - \log \Big(c_p(1)\exp(\bml_c^T\bml_p')\Big) \\
% = &\ \log c_p(1) + \bml_c^T\bml_p - \log c_p(1) - \bml_c^T\bml_p' \\
 = &\ \ \bml_c^T\bml_p - \bml_c^T\bml_p'.
\end{split}
\label{eqn:elim}
\end{equation}
Therefore,
\begin{equation}
  \mathcal{J}_L = \sum_{l_c \in \mathcal{T}\backslash \{{\rm ROOT}\}} \sum_{l_p' \in \mathcal{T} \backslash \{l_p, l_c\}} \min\Big(0, \bml_c^T\bml_p - \bml_c^T\bml_p' - \gamma_L \Big).
\label{eqn:jl}
\end{equation}

\noindent \textbf{Objective for Metadata and Corpus.} Given the label and metadata information of a document $d$, the log-likelihood of observing the document and its word sequence is 
\begin{equation}
 \log p(d, w_{1:n}|l_d, \mathcal{M}_d)
 = \log \Big(p(d|l_d, \mathcal{M}_d) \prod_{i=1}^n p(w_i|d) \Big).
\end{equation}
According to Eqs. (\ref{eqn:ld2}) and (\ref{eqn:dw1}), it can be written as
\begin{equation}
\sum_{z \in \mathcal{O}_d} \log f_{\rm vMF}(\bmd|\bmz,1) + \sum_{i=1}^n \log f_{\rm vMF}(\bmw_i|\bmd,1) + {\rm const}.
\end{equation}
Again, we adopt the margin-based ranking objective to push the negative pairs distant enough from the positive pairs $(z, d)$ and $(d, w_i)$. Formally,
\begin{equation}
\begin{split}
\mathcal{J}_{MC} = &\sum_{d \in \mathcal{D}} \sum_{d'\in \mathcal{D}\backslash\{d\}} \bigg( \\ &\sum_{z\in \mathcal{O}_d} \min\Big(0, \log f_{\rm vMF}(\bmd|\bmz,1) - \log f_{\rm vMF}(\bmd'|\bmz,1) - \gamma_M \Big) + \\
&\sum_{i=1}^n \min\Big(0, \log f_{\rm vMF}(\bmw_i|\bmd,1) - \log f_{\rm vMF}(\bmw_i|\bmd',1) - \gamma_C \Big) \bigg).
\end{split}
\end{equation}
Following the derivation in Eq. (\ref{eqn:elim}), we have
\begin{equation}
\begin{split}
\mathcal{J}_{MC} = \sum_{d \in \mathcal{D}} \sum_{d'\in \mathcal{D}\backslash\{d\}} \bigg( & \sum_{z\in \mathcal{O}_d} \min\Big(0, \bmd^T\bmz - \bmd'^T\bmz - \gamma_M \Big) + \\
& \sum_{i=1}^n \min\Big(0, \bmw_i^T\bmd - \bmw_i^T\bmd' - \gamma_C \Big) \bigg).
\end{split}
\label{eqn:jmc}
\end{equation}

To summarize, our joint representation learning step can be formulated as the following optimization problem:
\begin{equation}
\begin{gathered}
    \max \mathcal{J} = \mathcal{J}_L + \mathcal{J}_{MC}, \\
    {\rm s.t.\ \ \ \ } ||\bmth||_2 = 1,\ \ \ \ \forall \theta \in \mathcal{T} \cup \mathcal{M}\cup \mathcal{D} \cup \mathcal{V}.
\end{gathered}
\label{eqn:obj}
\end{equation}
Here, $\mathcal{M}$ is the set of metadata instances appearing in the dataset, and $\mathcal{V}$ is the vocabulary (i.e., the set of words).

\vspace{1mm}

\noindent \textbf{Optimization.} Given the constraint that all embedding vectors live on a sphere, directly performing stochastic gradient descent in the Euclidean space is not an effective way to optimize the objective. Instead, following recent studies on hyperbolic embedding \cite{nickel2017poincare,nickel2018learning} and spherical embedding \cite{wilson2014spherical,meng2019spherical}, we apply the Riemannian gradient method \cite{bonnabel2013stochastic} to learn the embeddings.

According to \cite{meng2019spherical}, on a unit sphere, the relationship between the Riemannian gradient $\nabla^R$ and the Euclidean gradient $\nabla^E$ is
\begin{equation}
   \nabla^R \mathcal{J}(\bmth) = ({\bm I} - \bmth\bmth^T)\nabla^E \mathcal{J}(\bmth).
\label{eqn:re}
\end{equation}
Based on Eqs. (\ref{eqn:jl}) and (\ref{eqn:jmc}), the Euclidean gradient of each embedding vector is easy to compute. For example, given a parent-child label pair $(l_p, l_c)$, we have
\begin{equation}
\begin{gathered}
\nabla^E \mathcal{J}_L(\bml_c) = \sum_{l_p' \in \mathcal{T} \backslash \{l_p, l_c\}} {\bf 1}(\bml_c^T\bml_p - \bml_c^T\bml_p' < \gamma_L)\cdot (\bml_p - \bml_p'), \\
\nabla^E \mathcal{J}_L(\bml_p) = \sum_{l_p' \in \mathcal{T} \backslash \{l_p, l_c\}} {\bf 1}(\bml_c^T\bml_p - \bml_c^T\bml_p' < \gamma_L)\cdot \bml_c, \\
\nabla^E \mathcal{J}_L(\bml_p') = {\bf 1}(\bml_c^T\bml_p - \bml_c^T\bml_p' < \gamma_L)\cdot (-\bml_c),\ \ \ \forall \bml_p' \in \mathcal{T} \backslash \{l_p, l_c\},
\end{gathered}
\end{equation}
where ${\bf 1}(\cdot)$ is the indicator function. Then, following Eq. (\ref{eqn:re}), we can obtain the Riemannian gradient $\nabla^R \mathcal{J}(\bml_c)$, $\nabla^R \mathcal{J}(\bml_p)$ and $\nabla^R \mathcal{J}(\bml_p')$. The gradient of other embedding vectors can be calculated in a similar way.
After knowing the Riemannian gradient, we update the parameters in the following form \cite{bonnabel2013stochastic,meng2019spherical}:
%along the gradient direction. Following \cite{meng2019spherical}, the update rule is as follows:
\begin{equation}
\bmth_{t+1} \leftarrow \frac{\bmth_t+\alpha_t \nabla^R \mathcal{J}(\bmth_t)}{||\bmth_t+\alpha_t \nabla^R \mathcal{J}(\bmth_t)||_2},
\end{equation}
where $\alpha_t$ is the learning rate at step $t$. Intuitively, this update can be viewed as an addition along the Riemannian gradient direction followed by a projection onto the unit sphere.

Since computing the gradient requires enumerating all labels or documents, in our actual computation, we adopt the negative sampling strategy \cite{mikolov2013distributed,bordes2013translating} to accelerate this process. To be specific, for each positive pair $(l_p, l_c)$, $(z, d)$ or $(d, w_i)$, we sample 5 negative pairs to estimate the gradient.

\subsection{Hierarchical Data Augmentation}
\newlength{\textfloatsepsave} 
\setlength{\textfloatsepsave}{\textfloatsep} 
\setlength{\textfloatsep}{1em}
\begin{algorithm}[!t]
\small
\caption{Hierarchical Data Augmentation}
\label{alg:gen}
\begin{algorithmic}[1]
    \STATE \textit{Input:} embedding vectors ($\bml, \bmd, \bmw$); \# synthesized docs per category ($\beta$)
    \STATE Generate a bottom-up traverse order of $\mathcal{T}$
	\FOR {$l$ in the traverse order}
	    \IF {$l$ is a leaf category}
	        \STATE $\mathcal{D}^*_l = \emptyset$
	        \FOR {$i=1$ to $\beta$}
	            \STATE Generate $\bmd^*$ according to Eq. (\ref{eqn:gen1})
	            \STATE Generate $\bmw^*_1\bmw^*_2...\bmw^*_n$ according to Eq. (\ref{eqn:gen2})
	            \STATE $\mathcal{D}^*_l = \mathcal{D}^*_l \cup \{d^*\}$
	        \ENDFOR
	    \ELSE
	        \FOR {$x \in \texttt{Child}(l)$}
	            \STATE $\mathcal{D}^*_{x, sample} \leftarrow$ sample $\frac{\beta}{|\texttt{Child}(l)|}$ docs from $\mathcal{D}^*_x$
	        \ENDFOR
	        \STATE $\mathcal{D}^*_l = \cup_{x \in \texttt{Child}(l)} \mathcal{D}^*_{x, sample}$
	    \ENDIF
	\ENDFOR
	\STATE \textit{Output:} $\{\mathcal{D}^*_l : l \in \mathcal{T}\}$
\end{algorithmic}
\end{algorithm}

Now we proceed to tackle the limited training data challenge. After representation learning, we have obtained the embedding vectors of all observed labels, metadata instances, documents and words. Using these embeddings, we can generate \textit{synthesized} training documents for each category. Formally, given a category $l$, we aim to create a document set $\mathcal{D}_l^*$ to augment its original training set $\mathcal{D}_l$. To implement this idea, we seek guidance from our proposed generative process once again.

\vspace{1mm}

\noindent \textbf{Data Augmentation for a Leaf Category.} Recall Figure \ref{fig:story}(a). If $l$ is a leaf category, it will directly participate in the generation of document embeddings according to Eq. (\ref{eqn:ld2}). If we denote a synthesized document as $d^*$ (to distinguish it from a ``real'' document $d$), the generation probability will become
\begin{equation}
    p(d^*|l, \mathcal{M}_{d^*}) \propto \prod_{z \in \mathcal{O}_{d^*}} f_{\rm vMF}(\bmd^*|\bmz,\kappa).
\label{eqn:gen0}
\end{equation}
Note that we do not have any metadata information of the synthesized document. Therefore, the only observable variable here in $\{l\}\cup \mathcal{M}_{d^*}$ is the label $l$. Also, it is not necessary to generate documents associated with a certain metadata instance because our goal is to predict the label, instead of the metadata information, of a document. Substituting $\mathcal{O}_{d^*} = \{l\}$ into Eq. (\ref{eqn:gen0}), we have
\begin{equation}
    p(d^*|l, \mathcal{M}_{d^*}) = f_{\rm vMF}(\bmd^*|\bml,\kappa).
    \label{eqn:gen1}
\end{equation}

After getting the document embedding $\bmd^*$, we continue our generative process to create a sequence of words describing $d^*$. In principle, we should follow Eq. (\ref{eqn:dw1}) and sample words from $f_{\rm vMF}(\cdot|\bmd^*, 1)$. However, in practice, we expect that the generated word embedding $\bmw^*$ can be mapped back to a concrete word that has occurred in our vocabulary $\mathcal{V}$. In \cite{kumar2018mises}, Kumar and Tsvetkov propose a way that first samples an vector $\bm e$ from a continuous distribution, and then find the nearest neighbor of $\bm e$ in the embedding space among all $w \in \mathcal{V}$. In contrast, we directly require the generated word to reside in the neighborhood of $\bmd^*$ in our joint embedding space. In this way, Eq. (\ref{eqn:dw1}) degenerates to a discrete softmax function:
\begin{equation}
    p(w^*|d^*) = \frac{f_{\rm vMF}(\bmw^*|\bmd^*,1)}{\sum_{w\in \mathcal{V}_{N}(d^*)} f_{\rm vMF}(\bmw|\bmd^*,1)} = \frac{\exp(\bmw^{* T}\bmd^*)}{\sum_{w\in \mathcal{V}_{N}(d^*)} \exp(\bmw^{T}\bmd^*)},
    \label{eqn:gen2}
\end{equation}
where $\mathcal{V}_{N}(d)$ is the set of top-$N$ nearest neighbors of $\bmd^*$ in the embedding space among all $w \in \mathcal{V}$. Using Eq. (\ref{eqn:gen2}) repeatedly, we can obtain a sequence of words $w^*_1w^*_2...w^*_n$ as the content of $d^*$.

\vspace{1mm}

\noindent \textbf{Data Augmentation for a Non-Leaf Category.} In our hierarchical generative process, a non-leaf label does not directly generate documents. Instead, it determines the embeddings of its children, and the child categories further determine its lower-level descendants. Formally, given a non-leaf label $l$, let $\mathcal{T}_l \subseteq \mathcal{T}$ be the subtree with root $l$ and $\texttt{Leaf}(\mathcal{T}_l)$ be the leaf descendants of $l$. Then,
\begin{equation}
p(\mathcal{D}^*_l, \mathcal{T}_l|l) = p(\mathcal{T}_l|l) \cdot \prod_{x \in \texttt{Leaf}(\mathcal{T}_l)}p(\mathcal{D}^*_x|x).
\label{eqn:gen3}
\end{equation}
After the joint representation learning step, all label embeddings have been fixed. Therefore, for any parent-child pair $(l_p, l_c)$, the probability $p(l_c|l_p) = c_p(1)\exp(\bml_c^T\bml_p)$ is fixed. Thus,
\begin{equation}
    p(T_l|l) = p(\texttt{Child}(l)|l) \cdot \prod_{x\in \texttt{Child}(l)} p(\texttt{Child}(x)|x) \cdot ... = {\rm const.}
\label{eqn:gen4}
\end{equation}
Here, $\texttt{Child}(l)$ is the set of $l$'s children. Putting Eqs. (\ref{eqn:gen3}) and (\ref{eqn:gen4}) together, we know that the generation of $\mathcal{D}^*_l$ solely depends on $\texttt{Leaf}(\mathcal{T}_l)$ in the current data augmentation step. That being said, for a leaf category $x \in \texttt{Leaf}(\mathcal{T}_l)$, the documents $\mathcal{D}^*_x$ sampled from $p(\mathcal{D}^*_x|x)$ can also be viewed as the documents $\mathcal{D}^*_l$ sampled from $p(\mathcal{D}^*_l, \mathcal{T}_l|l)$. This is intuitive because training samples of a leaf category are naturally training samples of its ancestor categories. We have discussed how to generate synthesized documents for a leaf category. Therefore, we can directly have
\begin{equation}
   \mathcal{D}^*_l = \cup_{x \in \texttt{Leaf}(\mathcal{T}_l)} \mathcal{D}^*_x.
\end{equation}
If we adopt a bottom-up order to generate $\mathcal{D}^*_l$, this can also be written in a recursive way:
\begin{equation}
   \mathcal{D}^*_l = \cup_{x \in \texttt{Child}(l)} \mathcal{D}^*_x.
 \label{eqn:agg1}
\end{equation}

In practice, some non-leaf categories have more children than their siblings have. Thus, if we directly apply Eq. (\ref{eqn:agg1}), they will have more synthesized training data. To avoid class imbalance, we fix $|\mathcal{D}^*_l| = \beta = \text{const}$ for any $l \in \mathcal{T}$. When applying Eq. (\ref{eqn:agg1}), we sample $\frac{\beta}{|\texttt{Child}(l)|}$ documents from $\mathcal{D}^*_x$ and put them into $\mathcal{D}^*_l$. Algorithm \ref{alg:gen} completely summarizes our data augmentation process.

\subsection{Hierarchical Classifier Training}
After hierarchical data augmentation, every (leaf or non-leaf) category $l$ has a set of synthesized training documents $\mathcal{D}^*_l$ and a set of real training data $\mathcal{D}_l$. (If $l$ is a non-leaf category, we follow the idea of creating $\mathcal{D}^*_l$ and let $\mathcal{D}_l = \cup_{x\in \texttt{Leaf}(\mathcal{T}_l)} \mathcal{D}_x$.) Each training document $d \in \mathcal{D}_l \cup \mathcal{D}^*_l$ has a sequence of words, whose embedding vectors are already learned in joint representation learning.

Now, based on the augmented training data, we need to train a hierarchical text classifier. Hierarchical text classification has been extensively studied for decades with many effective models proposed (see Section \ref{sec:related} for a detailed discussion). The main goal of this paper is to tackle signal heterogeneity and supervision scarcity instead of proposing a novel neural architecture for hierarchical classification. Therefore, we simply adopt the top-down training strategy \cite{koller1997hierarchically,dumais2000hierarchical,liu2005support} that classifies documents at the top layer and then propagates the results to the next layer until the leaves. For each non-leaf category $l \in \mathcal{T}$, we need to train a flat classifier that assigns documents to its child classes $\texttt{Child}(l)$ for more fine-grained predictions. We adopt Kim-CNN \cite{kim2014convolutional}, a popular convolutional architecture, as our flat text classifier. The embeddings $\{\bmw:w\in \mathcal{V}\}$ trained in the joint representation learning step are used as initialization embeddings of the input layer.

The model adopted here can be easily improved by replacing Kim-CNN with a more advanced flat classifier (e.g., \cite{yang2016hierarchical,wang2017hybrid}) or by adding a hierarchical regularization (e.g., \cite{gopal2013recursive,gopal2015hierarchical,peng2018large}). However, since we would like to demonstrate the contribution of our embedding and augmentation modules when comparing with baselines, we keep the classifier as simple as possible.

\setlength{\textfloatsep}{\textfloatsepsave}

\section{Experiments}
\subsection{Experimental Setup}
\begin{table}[t]
\caption{Dataset Statistics.}
\vspace{-1em}
\scalebox{0.94}{
	\begin{tabular}{c|ccccc}
		\hline
		Dataset     & \#Docs    & \#Classes   & \#Leaves  & \#Training & \#Testing \\
		\hline
		GitHub      & 1,596        & 18    & 14       & 70     & 1,526     \\
		ArXiv       & 26,400       & 94   & 88        & 440      & 25,960  \\
		Amazon      & 147,000      & 166  & 147       & 735    & 146,265  \\
		\hline
	\end{tabular}
}
\vspace{-1em}
\label{tab:data}
\end{table}

\noindent \textbf{Datasets.} We use three datasets from different domains.\footnote{Our code and datasets are available at \\ \url{https://github.com/yuzhimanhua/HIMECat}.} The statistics of the datasets can be found in Table \ref{tab:data}.

\begin{itemize}[leftmargin=*]
\item \textbf{GitHub \cite{zhang2019higitclass}.} This dataset is collected by paperswithcode.com. It contains 1,596 GitHub repositories implementing state-of-the-art machine learning algorithms for different tasks (e.g., object detection, question answering, speech synthesis, etc.). The tasks form a taxonomy which is viewed as the label hierarchy.
 
\item \textbf{ArXiv.} This dataset is crawled from arXiv.org. There are 5 level-1 categories (i.e., cs, math, physics, q-bio, and q-fin) and 88 level-2 categories. For each category, we randomly sample 300 papers to constitute the dataset.

\item \textbf{Amazon \cite{mcauley2013hidden}.} This dataset is a large collection of Amazon product reviews. The label of each review is its product category (e.g., automotive, car care, beauty, skin care, etc.). These categories are organized into a catalog taxonomy \cite{mao2020octet}. We select 147 large leaf categories and sample 1,000 reviews for each of them.
\end{itemize}
According to our weakly supervised setting, we use 5 documents per leaf category for training and all the other documents for testing.

\vspace{1mm}

\noindent \textbf{Baseline Methods.} We evaluate the performance of \textsc{HiMeCat} against the following classification/embedding algorithms:

\begin{itemize}[leftmargin=*]
\item \textbf{HierSVM \cite{dumais2000hierarchical}} is a supervised hierarchical classification method. It adopts a top-down classification strategy and trains an SVM for each non-leaf node to distinguish its child categories.
\item \textbf{WeSHClass \cite{meng2019weakly}} is a weakly supervised hierarchical classification method. It adopts LSTM to generate data for pre-training and iteratively refines results based on the hierarchy.
\item \textbf{PCEM \cite{xiao2019efficient}} is a weakly supervised hierarchical classification method. It proposes a path-cost sensitive hierarchical classifier and applies an EM technique to utilize the unlabeled data.
\item \textbf{HiGitClass \cite{zhang2019higitclass}} is a weakly supervised hierarchical classification method. It uses heterogeneous network embedding to encode relationships between labels, text and metadata and then leverages WeSHClass to train a hierarchical classifier. 
\item \textbf{MetaCat \cite{zhang2020minimally}} is a weakly supervised flat classification method. It leverages metadata information but cannot utilize the label hierarchy. We use it to directly classify documents to the leaf layer and then infer higher-level labels using the hierarchy.
\item \textbf{Metapath2vec \cite{dong2017metapath2vec}} is a metadata-aware embedding method. It models the proximity between heterogeneous elements through meta-path guided random walks.
\item \textbf{Poincar{\'e} \cite{nickel2017poincare}} is a hierarchy-aware embedding method. It preserves the tree structure of embedded elements by putting them into a hyperbolic space.
\item \textbf{Pretrained BERT \cite{devlin2019bert}} is a benchmark language model that provides contextualized word representations. For text classification, following \cite{devlin2019bert}, we add a classification layer upon the [CLS] token of the last transformer layer and fine-tune the model using labeled documents. After predicting the leaf category of a document, we infer its higher-level labels using the hierarchy.
\end{itemize}

For Metapath2vec and Poincar{\'e}, inspired by our hierarchical generative process, we construct a graph with edges $(l_p, l_c)$, $(l, d)$, $(m, d)$ and $(d, w)$ and perform node embedding on the graph. (We use four meta-paths $l$-$l$, $d$-$l$-$d$, $d$-$m$-$d$ and $d$-$w$-$d$ for Metapath2vec.) Then, to apply these two baselines in weakly supervised hierarchical classification, we use the learned embeddings to train a WeSHClass classifier (by replacing its word2vec embedding module). Since some baselines only utilize text information, for each document, we append its metadata instances to the end of its text sequence so that they can exploit these signals.

\vspace{1mm}

\noindent \textbf{Parameters.} Embedding dimension $p=100$ for all compared methods except BERT (whose base model is 768-dimensional). Margin hyperparameter $\gamma_L=\gamma_M=\gamma_C=0.2$. Document-specific vocabulary size $N = 50$. Number of synthesized documents per category $\beta = 500$. For the Kim-CNN classifier, we use one convolutional layer whose filter widths are 2, 3, 4, 5 with 20 feature maps each.

\vspace{1mm}

\noindent \textbf{Evaluation Metrics.} We use Micro and Macro F1 scores as evaluation metrics. Denote $TP_l$, $FP_l$ and, $FN_l$ as the instance numbers of true positive, false positive and false negative for a category $l \in \mathcal{T}$. The \underline{\textbf{Leaf Micro F1}} is defined as $\frac{2PR}{P+R}$, where $P=\frac{\sum_{l \in \texttt{Leaf}(\mathcal{T})} TP_l}{\sum_{l \in \texttt{Leaf}(\mathcal{T})} (TP_l+FP_l)}$ and $R=\frac{\sum_{l \in \texttt{Leaf}(\mathcal{T})} TP_l}{\sum_{l \in \texttt{Leaf}(\mathcal{T})} (TP_l+FN_l)}$. The \underline{\textbf{Leaf Macro}} \underline{\textbf{F1}} is defined as $\frac{1}{|\texttt{Leaf}(\mathcal{T})|}\sum_{l \in \texttt{Leaf}(\mathcal{T})}\frac{2P_lR_l}{P_l+R_l}$, where $P_l = \frac{TP_l}{TP_l+FP_l}$ and $R_l = \frac{TP_l}{TP_l+FN_l}$. Following previous works \cite{meng2019weakly,xiao2019efficient}, we also calculate the \underline{\textbf{Overall Micro/Macro F1}}, which is defined accordingly on $\mathcal{T}\backslash \{\rm ROOT\}$ instead of $\texttt{Leaf}(\mathcal{T})$.

\subsection{Performance Comparison}
\begin{table*}[t]
\caption{\{Leaf, Overall\}$\times$\{Micro, Macro\} F1 scores of compared algorithms on the three datasets. *: significantly worse than \textsc{HiMeCat} (p-value $< 0.05$). **: significantly worse than \textsc{HiMeCat} (p-value $< 0.01$).}
\vspace{-1em}
\scalebox{0.91}{
\begin{tabular}{c|cccc|cccc|cccc}
\hline
             & \multicolumn{4}{c|}{GitHub}                                           & \multicolumn{4}{c|}{ArXiv}                                            & \multicolumn{4}{c}{Amazon}                                            \\ \cline{2-13} 
             & \multicolumn{2}{c}{Leaf}          & \multicolumn{2}{c|}{Overall}      & \multicolumn{2}{c}{Leaf}          & \multicolumn{2}{c|}{Overall}      & \multicolumn{2}{c}{Leaf}          & \multicolumn{2}{c}{Overall}       \\ \cline{2-13}
             & Micro           & Macro           & Micro           & Macro           & Micro           & Macro           & Micro           & Macro           & Micro           & Macro           & Micro           & Macro           \\ \hline
HierSVM \cite{dumais2000hierarchical}      & 0.1861**          & 0.1388**          & 0.4862**          & 0.2457**          & 0.0538**          & 0.0460**          & 0.4066**          & 0.0750**          & 0.0248**          & 0.0217**          & 0.2218**          & 0.0494**          \\
WeSHClass \cite{meng2019weakly}   & 0.1727**          & 0.1559**          & 0.3332**          & 0.1924**          & 0.0604**          & 0.0602**          & 0.3077**          & 0.0797**          & 0.0483**          & 0.0500**          & 0.1234**          & 0.0640**          \\
PCEM \cite{xiao2019efficient}        & 0.2519**          & 0.1234**          & 0.5299*          & 0.1786**          & 0.1090**          & 0.0717**          & 0.4440          & 0.0963**          & 0.0675**          & 0.0439**          & 0.2189**          & 0.0659**          \\
HiGitClass \cite{zhang2019higitclass}   & 0.3984          & 0.3902*          & 0.5073**          & 0.4084**          & 0.1738**          & 0.1656**          & 0.3928**          & 0.1880**          & 0.0903**          & 0.0876**          & 0.1677**          & 0.1040**          \\ 
MetaCat \cite{zhang2020minimally}   &  0.3762**         &  0.3403**         &   0.5411*        &      0.3863**     &   0.0790**        &  0.0768**         & 0.3071**          &  0.0935**         &    0.1008**       &  0.0994**         &  0.1703**         &  0.1083**         \\ \hline
Metapath2vec \cite{dong2017metapath2vec} & 0.2814**          & 0.2805**          & 0.4592**          & 0.3212**          & 0.1360**          & 0.1344**          & 0.3419**          & 0.1534**          & 0.0669**          & 0.0666**          & 0.1334**          & 0.0800**          \\
Poincar{\'e} \cite{nickel2017poincare}    & 0.2750**          & 0.1980**          & 0.4302**          & 0.2218**          & 0.1336**          & 0.1296**          & 0.2995**          & 0.1454**          & 0.0645**          & 0.0607**          & 0.1202**          & 0.0739**          \\
BERT \cite{devlin2019bert}        &  0.2889**               &  0.2561**               &  0.4675**               &   0.3007**              &  0.1316**               &   0.0812**              &  0.4203**               &  0.1100**               &  0.0891**               &  0.0520**               &   0.2361**              &   0.0771**              \\ \hline
\textsc{HiMeCat}      & \textbf{0.4254} & \textbf{0.4209} & \textbf{0.5820} & \textbf{0.4535} & \textbf{0.2038} & \textbf{0.1938} & \textbf{0.4509} & \textbf{0.2191} & \textbf{0.1552} & \textbf{0.1553} & \textbf{0.2748} & \textbf{0.1770} \\ \hline
\end{tabular}
}
\vspace{-0.5em}
\label{tab:perform}
\end{table*}

\begin{table*}[t]
\centering
\begin{minipage}{0.565\linewidth}
\centering
\caption{Ablation analysis of the joint embedding module. --: the dataset does not have such metadata. * and **: the same as Table \ref{tab:perform}.}
\vspace{-1em}
\scalebox{0.915}{
\begin{tabular}{c|cc|cc|cc}
\hline
               & \multicolumn{2}{c|}{GitHub, Overall} & \multicolumn{2}{c|}{ArXiv, Overall} & \multicolumn{2}{c}{Amazon, Overall} \\ \cline{2-7} 
               & Micro             & Macro             & Micro             & Macro            & Micro             & Macro            \\ \hline
\textsc{HiMeCat}           & \textbf{0.5820}            & \textbf{0.4535}            & \textbf{0.4509}            & \textbf{0.2191}           & \textbf{0.2748}            & \textbf{0.1770}           \\ \hline
No-Hierarchy   & 0.5649*            & 0.4385            & 0.4353*            & 0.2007*           &      0.2640**            &   0.1658**               \\ \hline
No-Metadata    & 0.5262**            & 0.3949**            & 0.4208**            & 0.1951**           &   0.2196**                &    0.1117**              \\
No-User/Author & 0.5626*            & 0.4238**            & 0.4208**            & 0.1951**           &    0.2578**               &     0.1520**             \\
No-Tag         & 0.5649*            & 0.4263**            & --                 & --                & --                 & --                \\
No-Product     & --                 & --                 & --                 & --                &   0.2304**                &    0.1134**              \\ \hline
\end{tabular}
}
\label{tab:abl}
\end{minipage}
\hspace{0.5mm}
\begin{minipage}{0.42\linewidth}
\centering
\captionsetup{type=figure}
    \subfigure[Overall Micro F1]{
    \includegraphics[width=0.475\linewidth]{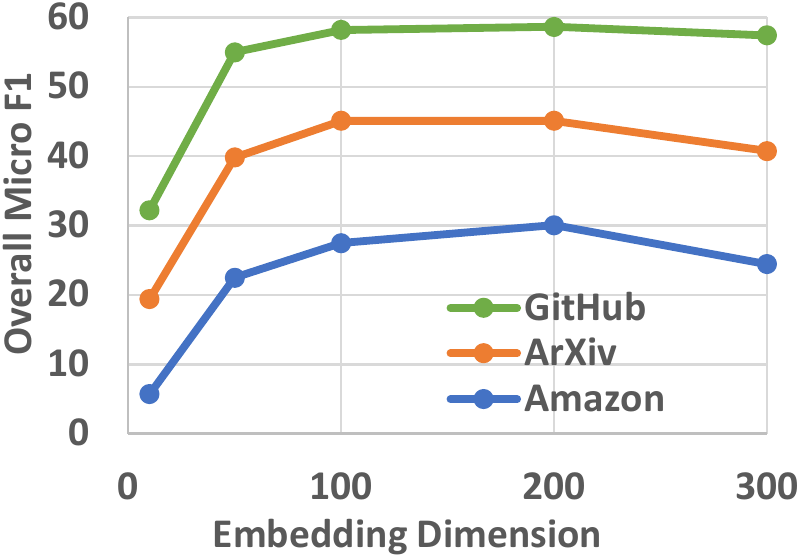}}
    \hspace{0mm}
    \subfigure[Overall Macro F1]{
    \includegraphics[width=0.475\linewidth]{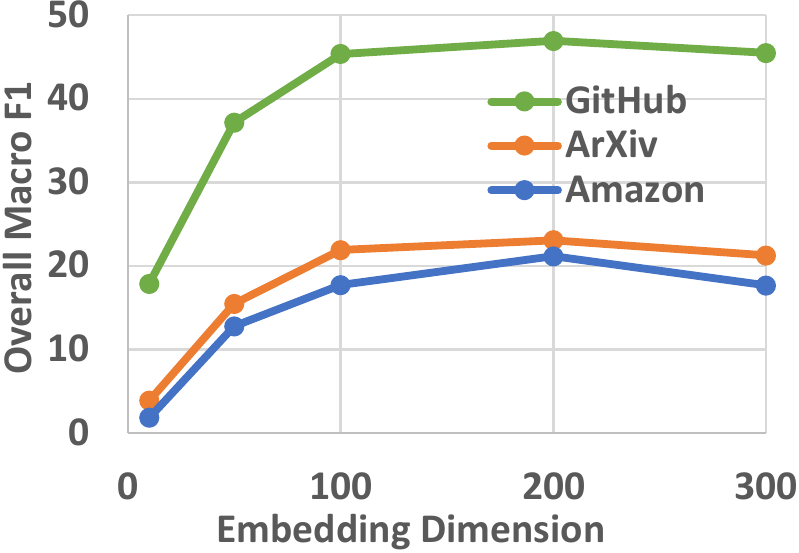}}
    \vspace{-1.5em}
\caption{Performance of \textsc{HiMeCat} with different embedding dimensions ($p$).}
\label{fig:dim}
\end{minipage}
\vspace{-1em}
\end{table*}

Table \ref{tab:perform} shows the categorization performance of compared algorithms on the three datasets. We repeat each experiment five times and report the average F1 scores. To measure statistical significance, we conduct a two-tailed paired t-test to compare \textsc{HiMeCat} and each baseline approach. The significance level (``p-value $< 0.01$'' or ``p-value $< 0.05$'') of each result is also marked in Table \ref{tab:perform}.

We observe that: (1) \textsc{HiMeCat} consistently outperforms the compared baseline methods on all three datasets. In most cases, the performance gap is statistically significant. The improvement of \textsc{HiMeCat} over the best baseline is 8.3\%, 13.0\% and 47.5\% on GitHub, ArXiv and Amazon, respectively. (2) For the three metadata-aware text categorization methods, \textsc{HiMeCat} performs better than HiGitClass and MetaCat, indicating the importance of considering label hierarchy and incorporating it into representation learning. (3) Although Metapath2vec and Poincar{\'e} are assumed to be strong tools for metadata and hierarchy embedding \textit{respectively}, they do not achieve competitive performance when we need to \textit{jointly} embed these signals. This observation also emphasizes the contribution of our joint representation learning module. (4) Despite the great success of BERT in many supervised NLP tasks, it does not perform well in our problem settings. There may be two possible reasons. First, directly using BERT for text classification does not consider metadata and hierarchy information. Second, the small set of training data may not be enough to fine-tune BERT.

\subsection{Ablation Study of Representation Learning}
Two key steps, representation learning and data augmentation, are proposed in our \textsc{HiMeCat} framework. Now we validate their contribution to the categorization performance through ablation analysis. We start from the representation learning module.

\vspace{1mm}

\noindent \textbf{Hierarchy and Metadata.} In our joint hierarchy-metadata-text representation learning module, to check whether the label hierarchy and various types of metadata are useful, we create the following ablation versions of \textsc{HiMeCat}.

\begin{itemize}[leftmargin=*]
\item \textbf{No-Hierarchy} does not consider parent-child relationships in representation learning. Formally, it ignores $\mathcal{J}_L$ in Eq. (\ref{eqn:obj}) and only maximizes $\mathcal{J}_{MC}$. 
\item \textbf{No-Metadata} does not consider metadata-document relationships in representation learning. Formally, in Eq. (\ref{eqn:jmc}), we will have $\mathcal{O}_d = \{l_d\}$ if $d$ is a training document and $\mathcal{O}_d = \emptyset$ otherwise. Moreover, instead of ignoring all metadata instances, we can overlook one specific type of metadata. This will yield ablations including \textbf{No-User/Author}, \textbf{No-Tag} and \textbf{No-Product}.
\end{itemize}

Table \ref{tab:abl} depicts the comparison between the full \textsc{HiMeCat} model and its ablations on the three datasets. (Due to space limitations, we only show Overall F1 scores. Similar observations can be drawn from Leaf F1 scores.) We find that: (1) The full model outperforms No-Hierarchy, which validates our claim that modeling label dependencies and correlation is beneficial to hierarchical categorization. Moreover, when the dataset has a larger label hierarchy, the gap between \textsc{HiMeCat} and No-Hierarchy is more statistically significant. The average improvement of \textsc{HiMeCat} over No-Hierarchy on the six metrics is 5.0\%. (2) The full model outperforms No-Metadata, which validates our claim that incorporating metadata signals is helpful. When studying different types of metadata separately, we find all of them play a positive role on our datasets. The contribution of metadata is larger than that of label hierarchy, possibly because metadata information is \textit{denser} than hierarchy information. That being said, each document has its own metadata, while the whole corpus only has one label hierarchy.

\vspace{1mm}

\noindent \textbf{Dimension.} The embedding dimension $p$ is an important parameter of the representation learning module. To check the sensitivity of $p$, we plot Overall F1 scores of \textsc{HiMeCat} with $p=10,50,100,200$ and $300$ in Figure \ref{fig:dim}. We can observe that: if $p$ is too small, the embedding vectors cannot sufficiently capture heterogeneous signals in the dataset; if $p$ is too large, we may face overfitting problems, especially under weak supervision. According to the results, setting $p$ between 100 and 200 is reasonable on our datasets.

\subsection{Ablation Study of Data Augmentation}
\begin{figure}[t]
    \centering
    % \subfigure[Leaf Micro F1]{
    % \includegraphics[width=0.48\linewidth]{syn1.pdf}}
    % \hspace{0mm}
    % \subfigure[Leaf Macro F1]{
    % \includegraphics[width=0.48\linewidth]{syn2.pdf}} \\
    % \vspace{-1em}
    \subfigure[Overall Micro F1]{
    \includegraphics[width=0.48\linewidth]{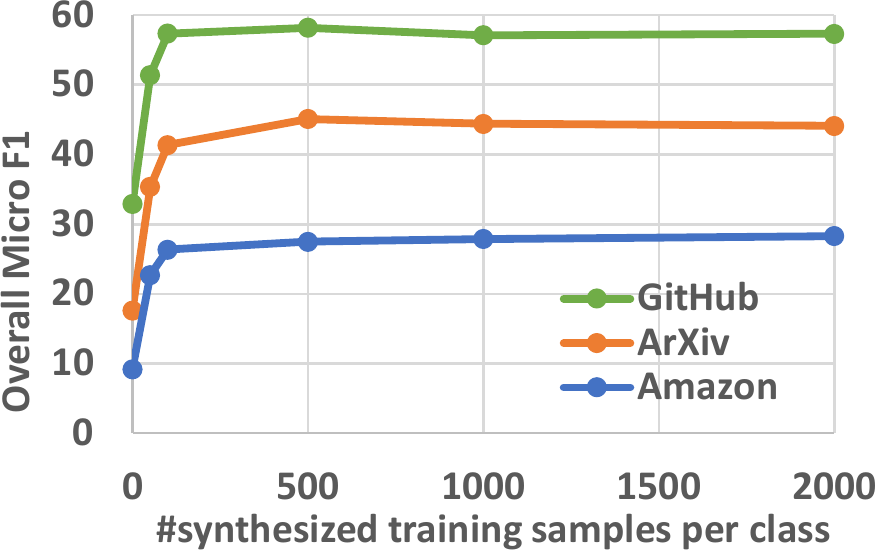}}
    \hspace{0mm}
    \subfigure[Overall Macro F1]{
    \includegraphics[width=0.48\linewidth]{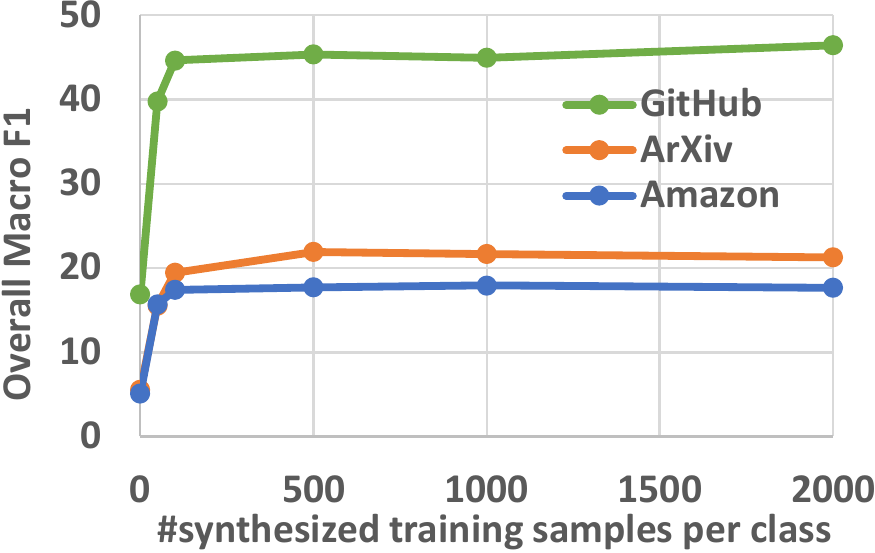}} \\
    \vspace{-1.5em}
    \caption{Performance of \textsc{HiMeCat} with different numbers of synthesized training samples per class ($\beta$).} 
    \vspace{-1em}
    \label{fig:syn}
\end{figure}
%Remove syn1/2 if space is limited

Now we proceed to the hierarchical data augmentation module. We need to answer two questions here: (1) Can data augmentation mitigate the label scarcity problem? (2) How many synthesized training samples should we create for each category? To investigate these two questions, we show the performance of \textsc{HiMeCat} with $\beta=0, 50, 100, 500, 1000$ and 2000 in Figure \ref{fig:syn}.

We observe that, when comparing $\beta = 500$ with $\beta = 0$ (i.e., we do not generate any synthesized training data and use real training data only), the performance boost is quite evident. For example, the average absolute improvement of Overall F1 scores on the three datasets is 21.4\%. Thus, we validate our claim that the hierarchical data augmentation module can help improve the classification accuracy under weak supervision. Meanwhile, the performance gap between $\beta = 2000$ with $\beta = 500$ is quite subtle. In fact, the average absolute gap of Overall F1 scores on the three datasets becomes 0.1\%. Also, too many synthesized training samples will make the training process inefficient. Seeking a balance, we believe setting $\beta=500$ is an appropriate choice (and we do set $\beta=500$ in previous experiments).

\subsection{Visualization of Tree Embedding}
\begin{figure}[t]
    \centering
    \subfigure[GitHub]{
    \includegraphics[width=0.48\linewidth]{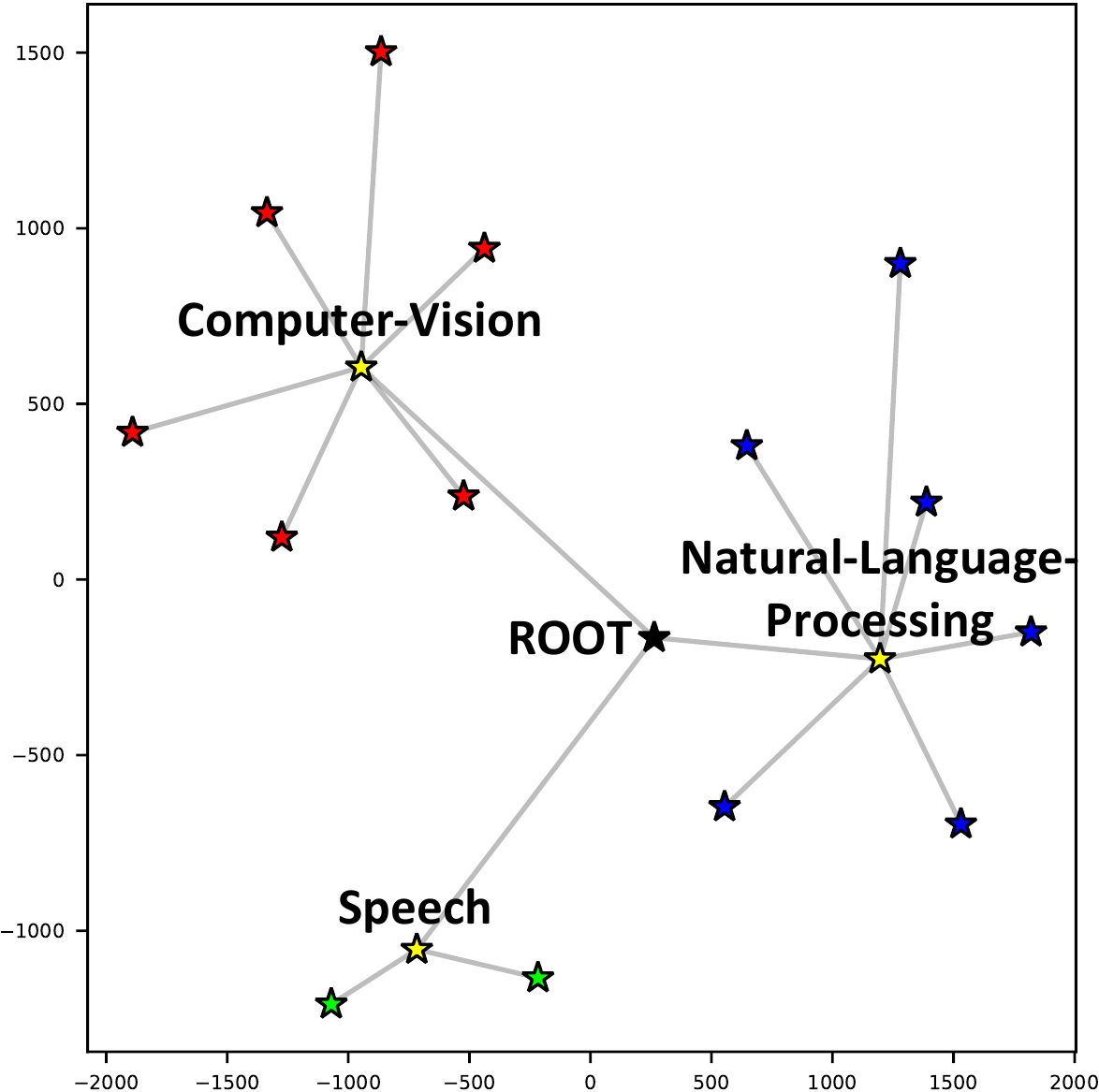}}
    \hspace{0mm}
    \subfigure[ArXiv]{
    \includegraphics[width=0.48\linewidth]{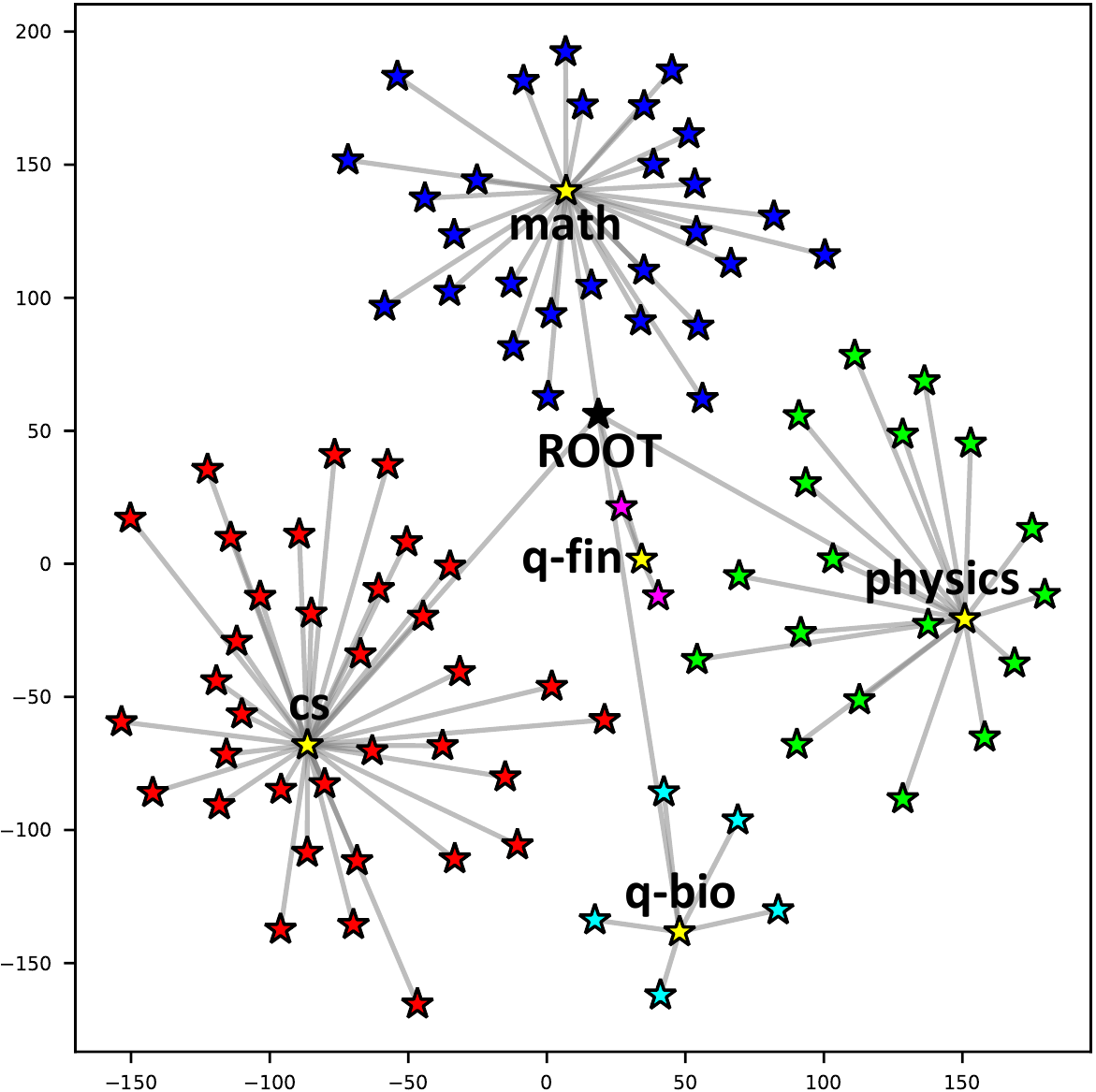}} \\
    \vspace{-1.5em}
    \caption{Tree Embedding Visualization.} 
    \vspace{-1em}
    \label{fig:vis}
\end{figure}

In order to show how the label hierarchy is modeled and how the labels are distributed in our joint embedding space, we plot label embeddings in Figure \ref{fig:vis} after reducing their dimensions using t-SNE \cite{maaten2008visualizing}. Labels are denoted as stars, and names of all level-1 categories and the root are shown. We have two observations: (1) The tree structure of the label hierarchy is well preserved in our embedding space. Child categories center around their parent categories. (2) In the ArXiv embedding space, ``q-fin'' (Quantitative Finance) is embedded near the root, and its children are embedded closer to ``math'' and ``physics'' than they are to ``cs'' and ``q-bio'' (Quantitative Biology). This is reasonable because quantitative finance is a quite interdisciplinary area and has many overlap with mathematics and physics. This observation shows that our joint representation learning module not only considers the label hierarchy but also captures semantics from text and/or metadata information.

\section{Related Work}
\label{sec:related}
\noindent \textbf{Hierarchical Text Classification.}
Many efforts have been put on how to utilizing the label hierarchy to improve text classification. For example, Dumais and Chen \cite{dumais2000hierarchical} and Liu et al. \cite{liu2005support} adopt a top-down training strategy and use SVMs to distinguish child categories of the same parent. In contrast, bottom-up classification \cite{bennett2009refined} backpropagates the labels from the leaves to the top layer. Gopal and Yang \cite{gopal2013recursive,gopal2015hierarchical} propose a recursive regularization framework encouraging the similarity between the child classifiers and the parent classifier. Peng et al. \cite{peng2018large} extend this regularization to deep graph-CNN models. 
%There are also effective and efficient frameworks on hierarchical extreme multi-label text classification, such as FastXML \cite{prabhu2014fastxml}, PfastreXML \cite{jain2016extreme}, SwiftXML \cite{prabhu2018extreme} and CRAFTML \cite{siblini2018craftml}. However, these methods are designed under fully supervised settings and rely heavily on the amount of human-labeled training data.
Wehrmann et al. \cite{wehrmann2018hierarchical} and Huang et al. \cite{huang2019hierarchical} combine the ideas of training a local classifier per level and optimizing the global classification results to mitigate exposure bias. The global structure of hierarchies is also used in many other models, such as meta-learning \cite{wu2019learning}, reinforcement learning \cite{mao2019hierarchical} and tree/graph based neural networks \cite{zhou2020hierarchy}.
However, these fully supervised methods rely heavily on the amount of human-labeled training data.
Under weakly supervised settings, hierarchical dataless classification \cite{song2014dataless} jointly embeds class labels and documents using Explicit Semantic Analysis; WeSHClass \cite{meng2019weakly} models topic semantics in the word2vec embedding space and applies a self-training scheme; 
%PCEM \cite{xiao2019efficient} introduces a path-cost sensitive classifier and applies an EM technique for semi-supervised learning; 
PCEM \cite{xiao2019efficient} introduces a path-cost sensitive classifier for semi-supervised hierarchical classification; 
HierCon \cite{li2019hiercon} projects documents and taxonomy categories into a common concept space and calculates their fine-grained similarity. However, these studies are not concerned with metadata information.
%, which \textsc{HiMeCat} makes extensive use of.

\vspace{1mm}

\noindent \textbf{Metadata-Aware Text Classification.} Several previous studies try to incorporate metadata information into a text classifier. For example, Ghani et al. \cite{ghani2001hypertext} use HTML meta tags to help hypertext classification; Steyvers et al. \cite{steyvers2004probabilistic} leverage author information in topic classification; Tang et al. \cite{tang2015learning} learn user and product representations for sentiment analysis; Zhang et al. \cite{zhang2017rate} employ user biography data for tweet localization. Chen et al. \cite{chen2020context} encode user mobility data and social network information for joint time and location prediction; Kim et al. \cite{kim2019categorical} propose a general framework to inject categorical metadata signal into a deep text classifier. However, these models assume a fully supervised setting. Zhang et al. \cite{zhang2019higitclass} present a weakly supervised hierarchical classification approach to classify GitHub repositories. 
%They use a heterogeneous network embedding method \cite{shang2016meta} to encode metadata information. 
Later, they also propose a flat metadata-aware text categorization framework \cite{zhang2020minimally}. Mekala et al. \cite{mekala2020meta} explore to incorporate metadata as additional supervision for text classification with seed words only. However, in these studies, the label hierarchy is not leveraged in the embedding space. 
%In contrast, \textsc{HiMeCat} proposes a novel representation learning module considering the dependency between parent and child labels.

\vspace{1mm}

\noindent \textbf{Tree and Metadata Embedding.} Recent studies on non-Euclidean embedding models, such as Poincar{\'e} \cite{nickel2017poincare}, Lorentz \cite{nickel2018learning}, hyperbolic cones \cite{ganea2018hyperbolic} and spherical tree embedding \cite{meng2020hierarchical}, consider to preserve a tree structure in the embedding space. Along another line of work, heterogeneous network embedding algorithms \cite{dong2017metapath2vec,fu2017hin2vec,yang2020heterogeneous} are widely used to learn representations of metadata. HHNE \cite{wang2019hyperbolic} further considers to perform heterogeneous network embedding in a hyperbolic space. 
Despite their \textit{respective} success in capturing hierarchy and metadata information, to the best of our knowledge, there lacks a framework which allows \textit{simultaneous} modeling of tree-structure dependencies, text semantics and metadata signals. 
%In our \textsc{HiMeCat} framework, we have a joint representation learning module facilitating this goal.

\section{Conclusions}
We present \textsc{HiMeCat}, an embedding-based generative framework for hierarchical metadata-aware document categorization under weak supervision. The framework is featured by a joint hierarchy-metadata-text representation learning module and a hierarchical data augmentation module. We propose a generative process in the spherical space to guide the design of both modules. Through experiments on three datasets from different domains, we show the superiority of \textsc{HiMeCat} towards competitive baselines in our task. We also conduct ablation studies to validate the contribution of our proposed embedding and augmentation modules. Interesting future work include: (1) discovering new categories from the unlabeled dataset and put them into the existing hierarchy and (2) integrating different forms of weak supervision (e.g., annotated documents and class-related keywords) in hierarchical text classification.

\begin{acks}
Research was sponsored in part by US DARPA KAIROS Program No. FA8750-19-2-1004 and SocialSim Program No. W911NF-17-C-0099, National Science Foundation IIS-19-56151, IIS-17-41317, IIS 17-04532, and IIS 16-18481, and DTRA HDTRA11810026. Any opinions, findings, and conclusions or recommendations expressed herein are those of the authors and should not be interpreted as necessarily representing the views, either expressed or implied, of DARPA or the U.S. Government. The U.S. Government is authorized to reproduce and distribute reprints for government purposes notwithstanding any copyright annotation hereon.
\end{acks}

\bibliography{wsdm}
\end{spacing}

\end{document}